%% file: main.tex
\documentclass{article}

\usepackage{microtype}
\usepackage{graphicx}
\usepackage{booktabs} % for professional tables
\usepackage{amsfonts} 
\usepackage[]{amsmath}
\usepackage{subcaption}

\newcommand{\ie}{\textit{i}.\textit{e}.}

\newcommand{\tabref}[1]{Table~\ref{#1}}

\newcommand{\figref}[1]{Figure~\ref{#1}}

\usepackage{amsthm}
\newtheorem{theorem}{Theorem}%[section]

\usepackage[pdftex,dvipsnames]{xcolor}  % Coloured text etc.
\usepackage[colorinlistoftodos,prependcaption,textsize=tiny]{todonotes}
\usepackage{xargs}                      % Use more than one optional parameter in a new commands
\newcommandx{\unsure}[2][1=]{\todo[linecolor=red,backgroundcolor=red!25,bordercolor=red,#1]{#2}}
\newcommandx{\change}[2][1=]{\todo[linecolor=blue,backgroundcolor=blue!25,bordercolor=blue,#1]{#2}}
\newcommandx{\info}[2][1=]{\todo[linecolor=OliveGreen,backgroundcolor=OliveGreen!25,bordercolor=OliveGreen,#1]{#2}}
\newcommandx{\improvement}[2][1=]{\todo[linecolor=Plum,backgroundcolor=Plum!25,bordercolor=Plum,#1]{#2}}
\newcommandx{\thiswillnotshow}[2][1=]{\todo[disable,#1]{#2}}
\usepackage{hyperref}

\usepackage[accepted]{icml2018}

\icmltitlerunning{Game-Theoretic Interpretability for Temporal Modeling}

\begin{document}

\twocolumn[
%\icmltitle{Game-theoretic Interpretability for Unconstrained Sequence Modeling}
\icmltitle{Game-Theoretic Interpretability for Temporal Modeling}



\icmlsetsymbol{equal}{*}

\begin{icmlauthorlist}
\icmlauthor{Guang-He Lee}{to}
\icmlauthor{David Alvarez-Melis}{to}
\icmlauthor{Tommi S. Jaakkola}{to}
\end{icmlauthorlist}

\icmlaffiliation{to}{MIT Computer Science and Artificial Intelligence Laboratory}

\icmlcorrespondingauthor{Guang-He Lee}{guanghe@mit.edu}
%\icmlcorrespondingauthor{Eee Pppp}{ep@eden.co.uk}

\icmlkeywords{Machine Learning, ICML}

\vskip 0.3in
]



\printAffiliationsAndNotice{}  % leave blank if no need to mention equal contribution

%%%%%%%%%%%%%%%%%%%%%%%%%%%%%%%%%%%%%%%%%%%%%%%%%%%%%%%%%%%%%%%%%%%%%%%%
% paper tex files

% Minor things are not mentioned in the paper:
% we use Ridge regression to fit the ARMA model to reduce the effect of multiple possible solutions from fitting small data (though it still induces the effect).
% The TV value are big in explicit setting and small in implicit setting are partially due to this.
% $\Lambda(\cdot)$ is learned by Cholesky factorization with strictly positive diagonals using soft-plus activation.
% We only sample 10\% of the sequence to impose the game in each batch for efficiency, but evaluation is based on all sequences.

\input{abstract2.tex}
\input{intro.tex}
\input{method2.tex}
\input{example2.tex}
\input{experiment.tex}
\input{conclusion.tex}

\bibliography{example_paper}
\bibliographystyle{icml2018}

%%%%%%%%%%%%%%%%%%%%%%%%%%%%%%%%%%%%%%%%%%%%%%%%%%%%%%%%%%%%%%%%%%%%%%%%

\end{document}

%% file: abstract2.tex
\begin{abstract}

% Tommi's version
Interpretability has arisen as a key desideratum of machine learning models alongside performance. Approaches so far have been primarily concerned with fixed dimensional inputs emphasizing feature relevance or selection. In contrast, we focus on temporal modeling and the problem of tailoring the {predictor}, functionally, towards an interpretable family. To this end, we propose a co-operative game between the \emph{predictor} and an \emph{explainer} without any a priori restrictions on the functional class of the predictor. The goal of the explainer is to highlight, locally, how well the predictor conforms to the chosen interpretable family of temporal models. Our co-operative game is setup asymmetrically in terms of information sets for efficiency reasons. We develop and illustrate the framework in the context of temporal sequence models with examples.

\end{abstract}

%% file: intro.tex
\section{Introduction}
\label{sec:intro}
State-of-the-art predictive models tend to be complex and involve a very large number of parameters. While the added complexity brings modeling flexibility, it comes at the cost of transparency or interpretability. This is particularly problematic when predictions feed into decision-critical applications such as medicine where understanding of the underlying phenomenon being modeled may be just as important as raw predictive power.

Previous approaches to interpretability have focused mostly on fixed-size data, such as scalar-feature datasets~\cite{lakkaraju2016interpretable} or image prediction tasks~\cite{selvaraju2016grad}. Recent methods do address the more challenging setting of sequential data \cite{Lei2016Rationalizing,arras2017relevant} in NLP tasks where the input is discrete. Interpretability for continuous temporal data has remained mostly unexplored~\cite{wu2018tree}. 

% David's version
%In this work, we propose a novel approach to model interpretability which naturally lends itself to (though is not limited to) time-series data. Our approach departs from other frameworks in the way it enforces interpretability. As opposed to classic interpretable models (\eg, classification trees), our framework operates on a wide range of prediction models, \ie, it does not restrict their architecture. On the other hand, differing from post-hoc \textit{extrinsic} explanation approaches, such as LIME \cite{Ribeiro2016Why} and related perturbation-based methods~\cite{alvarez2017causal}, the explanations are \textit{intrinsic} and trained \textit{in coordination} with the predictive model to be explained. Compared to existing intrinsic intepretable methods~\cite{al2017contextual}, where interpretation is estimated in a point-wise basis, our interpretation is intrinsically local.

% semi-intrinsic version
In this work, we propose a novel approach to model interpretability that is naturally tailored (though not limited to) time-series data. 
Our approach differs from interpretable models such as interpretation generators~\cite{al2017contextual,Lei2016Rationalizing} where the architecture or the function class is itself explicitly constrained towards interpretability, e.g., taking it to be the set of linear functions.
%constrained to give rise to interpretable predictions. 
We also differ from post-hoc explanations of black-box methods through local perturbations~\cite{Ribeiro2016Why, alvarez2017causal}. In contrast, we establish an intermediate regime, game-theoretic interpretability, where the predictor remains functionally complex but is encouraged during training to follow a locally interpretable form. 

\iffalse
 new notion of \emph{semi-intrinsic interpretability} that explanations are intrinsic to model training but extrinsic to model itself.
By such nature, the derived approach can operate on an unconstrained set of predictive/interpretable models, \ie, it does not restrict their architectures. 
Meanwhile, the explanations are \textit{intrinsic} in the objective and trained \textit{in coordination} with the predictive model to be explained. 
\fi

At the core of our approach is a game-theoretic characterization of interpretability. This is set up as a two-player co-operative game between \emph{predictor} and \emph{explainer}. The predictor remains a complex model whereas the explainer is chosen from a simple interpretable family. The players minimize asymmetric objectives that combine the prediction error and the discrepancy between the players. The resulting predictor is biased towards agreeing with a co-operative explainer. The co-operative game equilibrium is stable in contrast to  GANs~\cite{goodfellow2014generative}.

% Instead of generating post-hoc extrinsic explanations, we propose a game-theoretic characterization of interpretability, in which the 

% Things that have to be mentioned:
% \begin{itemize}
% \item Emphasize novelty of interpretability in sequences
% \item Related work:
% \item Importance of applications of interpretable temporal-modeling: medical, weather, etc.
% \begin{itemize}
% 	\item Tao's rationales \cite{Lei2016Rationalizing}
%     \item LRP for discrete sequences \cite{arras2017relevant}
%     \item Murdoch et al
%     \item Check if there's a paper for \url{github.com/emanuel-metzenthin/Lime-For-Time}
% \end{itemize}
% \item Non-sequential stuff but probably also need to me mentioned:
% \begin{itemize}
% 	\item LIME \cite{Ribeiro2016Why}
%     \item 
% \end{itemize}
% \end{itemize}
The main contributions of this work are as follows:
\begin{itemize}
	\vspace{-2mm}
	\itemsep0em
	\item A novel game-theoretic interpretability framework that can take a wide range of prediction models without architectural modifications.
    \item Accurate yet explainable predictors where the explainer is trained coordinately with the predictor to actively balance interpretability and accuracy.  
    % semi-intrinsic version
%    \item We establish a novel notion of semi-intrinsic interpretability that incorporates advantage from both intrinsic and extrinsic methods.
    \item Interpretable temporal models, validated through quantitative and qualitative experiments, including stock price prediction and a physical component modeling. 
\end{itemize}

\iffalse
Our main contribution is to establish a game-theoretic characterization of interpretability. 
The formulation only assumes the target model and the axiomatic set of interpretable models are trainable.
With the progression of the cooperative game, the functional behavior the target function is biased toward interpretable models.
The characterization also admits quantitative evaluation of predictive models. 
\fi

%% file: method2.tex
\section{Methodology}
\label{sec:method}

In this work, we learn a (complex) predictive target function $f\in \mathcal{F}: \mathcal{X} \to \mathcal{Y}$ together with a simpler function $g \in \mathcal{G}:\mathcal{X}\to\mathcal{Y}$ defined over an axiomatic class of interpretable models $\mathcal{G}$. We refer to functions $f$ and $g$ as the predictor and explainer, respectively, throughout the paper. Note that we need not make any assumptions on the function class $\mathcal{F}$, instead allowing a flexible class of predictors. In contrast,
the family of explainers $\mathcal{G}$ is explicitly and intentionally constrained such as the set of linear functions. As any $g\in \mathcal{G}$ is assumed to be interpretable, the family $\mathcal{G}$ does not typically itself suffice to capture the regularities in the data. We can therefore hope to encourage the predictor to remain close to such interpretable models only locally.

For expository purposes, we will develop the framework in a discrete time setting where the predictor maps  $x_t \in \mathcal{X}$ to $f(x_t) \in \mathcal{Y}$ for $t=\{1, 2, \dots\}$. 
The data are denoted as $\mathcal{D}=\{(x_1,y_1), (x_2,y_2), \dots\}$. 
%We then instantiate the predictor with deep sequence generative models and explainers with linear models.
We then instantiate the predictor with deep sequence generative models, and the explainers with linear models.

\subsection{Game-Theoretic Interpretability}

\begin{figure}	
	\begin{subfigure}{0.225\textwidth}
		\centering
		\includegraphics[width=1.15\linewidth]{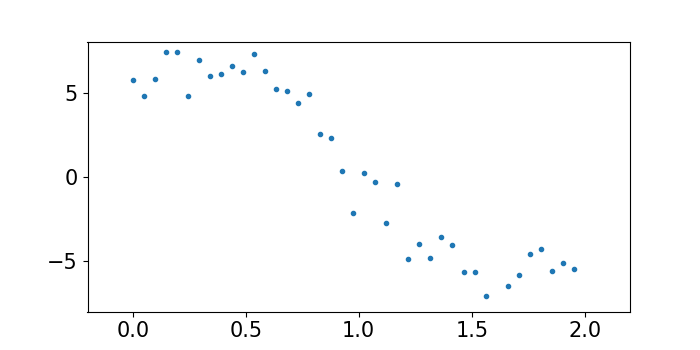}
		\caption{Neighborhood $\mathcal{B}_{\epsilon}(x_t=1)$}\label{fig:data}
	\end{subfigure}
	~
	\begin{subfigure}{0.225\textwidth}
		\centering
		\includegraphics[width=1.15\linewidth]{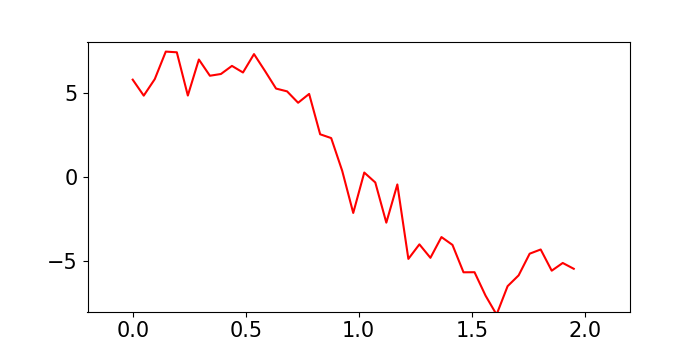}
		\caption{Piece-wise linear $f$}\label{fig:deep}
	\end{subfigure}

	\begin{subfigure}{0.225\textwidth}
		\centering
		\includegraphics[width=1.15\linewidth]{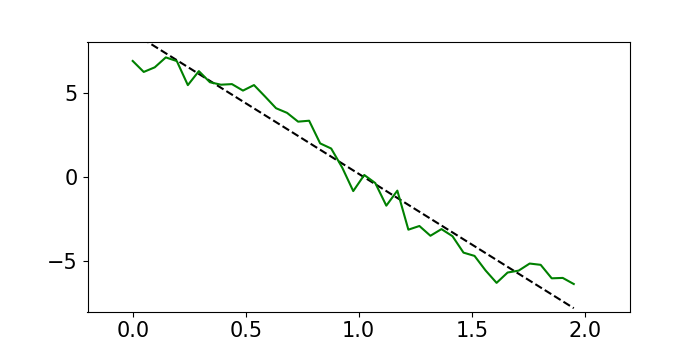}
		\caption{$g_{x_t} \in \mathcal{G}_{\text{Linear}}$}\label{fig:linear}
	\end{subfigure}
	~
	\begin{subfigure}{0.225\textwidth}
		\centering
		\includegraphics[width=1.15\linewidth]{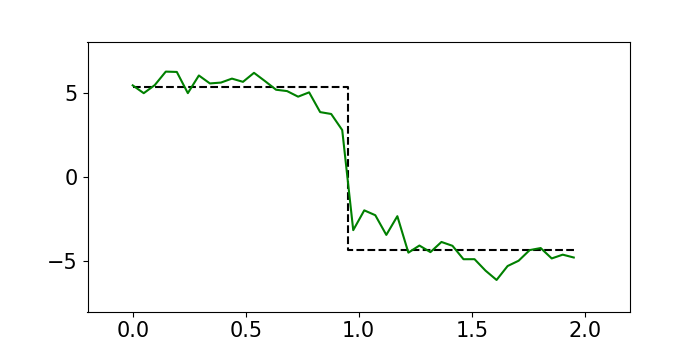}
		\caption{$g_{x_t} \in \mathcal{G}_{\text{Decision Stump}}$}\label{fig:stump}
	\end{subfigure}

	\caption{Examples of fitting a neighborhood $\mathcal{B}_{\epsilon}(x_t=1)$~(\ref{fig:data}) with piece-wise linear predictor~(\ref{fig:deep}). When playing with different families of explainers~(Figure \ref{fig:linear} and \ref{fig:stump}; dashed lines), the resulting predictor (in solid green) behaves differently although they admit the same prediction error (mean squared error = 1.026).}\label{fig:synthetic}
    \vspace{-1mm}
\end{figure}

There are many ways to use explainer functions $g \in \mathcal{G}$ to guide the predictor $f$ by means of discrepancy measures. However, since the explainer functions are inherently weak such as linear functions, we cannot expect that a reasonable predictor would be nearly linear globally. Instead, we can enforce this property only locally. To this end, we define \emph{local} interpretability by measuring how close $f$ is to a family $\mathcal{G}$ over a local neighborhood $\mathcal{B}_\epsilon(x_t)$ around an observed point $x_t$. One straightforward instantiation of such a neighborhood $\mathcal{B}_\epsilon({x_t})$ in temporal modeling will be simply a local window of points $\{x_{t-\epsilon},\dots, x_{t+\epsilon}\}$. Our resulting local discrepancy measure is
\begin{align}
\vspace{1mm}
& \min_{g \in \mathcal{G}} d_{x_t}(f, g)  = \min_{g \in \mathcal{G}} \frac{\sum_{x' \in \mathcal{B}_\epsilon(x_t)} d(f(x'), g(x'))}{|\mathcal{B}_\epsilon({x_t})|},
\label{eq:local_deviation}
\vspace{1mm}
\end{align}
where $d(\cdot, \cdot)$ is a deviation measurement.
The minimizing explainer function, $\hat g_{x_t}$, is indexed by the point $x_t$ around which it is estimated. Indeed, depending on the function $f$, the minimizing explainer can change from one neighborhood to another. If we view the minimization problem game-theoretically, $\hat g_{x_t}$ is the \emph{best response strategy} of the local explainer around $x_t$. 
  
The local discrepancy measure can be subsequently incorporated into an overall regularization problem for the predictor either symmetrically (shared objective) or asymmetrically (game-theoretic) where the goals differ between the predictor and the explainer.  

{\bf Symmetric criterion.} Assume that we are given a primal loss $\mathcal{L}(\cdot, \cdot)$ that is to be minimized for the problem of interest. The goal of the predictor is then to find $f$ that offers the best balance between the primal loss and local interpretability. 
Due to the symmetry between the two players, the full game can be collapsed into a single objective
%With the above characterization of interpretability, we can establish the complete game between predictor and local explainers to collaborate toward minimizing (\ref{eq:local_deviation}) on observed data $\mathcal{D}$. 
%Then the game is established as follows:
\begin{align}
%\min&_{f, g_{x_t}:(x_t, y_t)\in \mathcal{D}}  
\sum_{(x_t, y_t) \in \mathcal{D}}
\bigg[& \mathcal{L}(f(x_t), y_t) \nonumber\\
& + \frac{\lambda}{|\mathcal{B}_\epsilon(x_t)|} \sum_{x' \in \mathcal{B}_\epsilon(x_t)} d(f(x'), g_{x_t}(x'))\bigg]
%\label{eq:coop_game}
\end{align}
to be minimized with respect to both $f$ and $g_{x_t}$. Here, $\lambda$ is a hyper-parameter that we have to set.

To illustrate the above idea, we generate a synthetic dataset to show a \emph{neighborhood} in \figref{fig:data} with a perfect piece-wise linear predictor $f\in \mathcal{F_{\text{Piece-wise Linear}}}$ in \figref{fig:deep}. Clearly, $f$ does not agree with linear explainers within the neighborhood, despite its piece-wise linear nature. However, when we establish a game between $f$ and a linear explainer $g_{x_t=1} \in \mathcal{G}_{\text{Linear}}$ in \figref{fig:linear}, it admits lower functional deviation (and thus stronger linear interpretability). We also show in \figref{fig:stump} that different explainer family would induce different outcomes of the game.

%In (\ref{eq:coop_game}), the game between predictor function $f$ and explainer functions $g_{x_t}$ can be collapsed into a single objective, which makes it simple to understand the structure of cooperative game. 
%The symmetric structure in (\ref{eq:coop_game}) makes it simple to understand the objective of cooperative game, but there are some subtleties in the formulation to be addressed. 

{\bf Asymmetric Game.}
The symmetric criterion makes it simple to understand the overall co-operative objective, but solving it is inefficient computationally. Different possible sizes of the neighborhood $\mathcal{B}_\epsilon(x)$ (e.g., end-point boundary cases) makes it hard to parallelize optimization for $f$ (note that this does not hold for $g_{x_t}$), which is problematic when we require parallel training for neural networks. Also, since $f$ is reused many times across neighborhoods in the discrepancy measures, the value of $f$ at each $x_t$ may be subject to different functional regularization across the neighborhoods, which is undesirable. 

In principle, we would like to impose a uniform functional regularization for every $x_t$, where the regularizer is established on a local region $\mathcal{B}_\epsilon(x_t)$ basis. This new modeling framework leads to an asymmetric co-operative game, where the information sets are asymmetric between predictor $f$ and local explainers $g_{x_t}$. Accordingly, each local best response explainer $\hat g_{x_t}$ is minimized for local interpretability (\ref{eq:local_deviation}) within $\mathcal{B}_\epsilon(x_t)$, thus relying on $f$ values within this region. In contrast, the predictor $f$ only receives feedback in terms of resulting deviation at $x_t$ and thus only sees $\hat g_{x_t}(x_t)$. From the point of view of the predictor, the best response strategy is obtained by %minimizing 
\begin{align}
\vspace{-8mm}
\min_{f\in \mathcal{F}}\sum_{(x_t, y_t) \in \mathcal{D}} \mathcal{L}(f(x_t), y_t) + \lambda \cdot d(f(x_t), \hat g_{x_t}(x_t)).
\label{eq:asym_coop_game}
\vspace{-8mm}
\end{align}
%with respect to $f$.

{\bf Discussion.}
We can analyze the scenario when the loss and deviation are measured in squared error, the explainer is in constant family, and the predictor is non-parametric.
Both games induce a predictor that is equal to recursive convolutional average of $y_t$, where the decay rate in each recursion is the same $\frac{\lambda}{1+\lambda}$ for both games, but the convolutional kernel evolves twice faster in the symmetric game than in the asymmetric game. %, which indicates symmetric game would induce a smoother predictor.

The formulation involves a key trade-off between the size of the region where explanation should be simple and the overall accuracy achieved with the predictor. When the neighborhood is too small, local explainers become perfect, inducing no regularization on $f$. Thus the size of the region is a key parameter in our framework. Another subtlety is that solving (\ref{eq:local_deviation}) requires optimization over explainer family $\mathcal{G}$, where specific deviation and family choices matter for efficiency. For example, $L_2$ and affine family lead to linear regression with closed-form local explainers. Finally, a natural extension to solving (\ref{eq:local_deviation}) is to add regularizations. 

We remark that the setup of $f$ and $\mathcal{G}$ leaves considerable flexibility in tailoring the predictive model and the explanations to the application at hand --- which is not limited to temporal modeling. Indeed, the derivation is for temporal models but extends naturally to others as well. 

%% file: example2.tex
\section{Examples}
\label{sec:prob}

%To illustrate the asymmetric game, we can start from the most basic constant function family $\mathcal{G}_C$. Then for each $x_t$, $g_{x_t}$ is just the mean of $\{f(x_{t-\epsilon}),\cdots,f(x_{t+\epsilon})\}$, where the gradient should be disconnected from $g_x$ when updating $f$ due to game-theoretic setting. Similarly, for linear function family $\mathcal{G}_\text{Linear}$, we can also use a linear family by fitting a linear regression model based on data $\{(x_{t-\epsilon}, f(x_{t-\epsilon})),\cdots,(x_{t+\epsilon}, f(x_{t+\epsilon}))\}$

%\subsection{Conditional Generative Model}
{\bf Conditional Generative Model.} The basic idea of sequence modeling can be generalized to conditional sequence generation.
For example, given historical data of a stock's price, how will the price evolve over time?
Such mechanism allows us to inspect the temporal dynamics inside the problem of interest to assist in long-term decisions, while a conditional generation allows us to control the progression of generation with different settings of interest.

Formally, given an observation sequence $x_{1},\dots,$ $x_{t}\in\mathbb{R}^N$, the goal is to estimate the probability $p(x_{t+1},\dots,x_{T} | x_{1},\dots,x_{t})$ of future events $x_{t+1},\dots,$ $x_{T}\in \mathbb{R}^N$. For notational simplicity, we will use ${x}_{1:t}$ to denote the sequence of variables $x_1,\dots,x_t$.
A popular approach to estimate this conditional probability is to train a conditional model by maximum likelihood (ML)~\cite{van2016wavenet}. 
If we model the conditional distribution of $x_{i+1}$ given ${x}_{1:i}$ as a multivariate Gaussian distribution with mean $\mu(\cdot)$ and covariance $\Sigma(\cdot)$, we can define the asymmetric game on $\mu(\cdot)$ by minimizing
\begin{align}
\vspace{-2mm}
%\min_{\mu, \Sigma} 
\sum_{i=t}^{T-1} \bigg[& -\log \mathcal{N}(x_{i+1}; \mu({x}_{1:i}), \Sigma({x}_{1:i})) \nonumber \\
& + \lambda \; \|\mu({x}_{1:i}) - g_{{x}_{1:i}}({x}_{1:i})\|_2^2 \bigg],
\label{eq:implicit}
\end{align}
with respect to $\mu(\cdot)$ and $\Sigma(\cdot)$, both parametrized as recurrent neural networks.
For the explainer model $g_{{x}_{1:i}}(\cdot)$, we use the neighborhood data $\{({x}_{1:i-\epsilon}, \mu({x}_{1:i-\epsilon})),$ $\cdots,({x}_{1:i+\epsilon}, \mu({x}_{1:i+\epsilon}))\}$ to fit a $K$-order Markov autoregressive (AR) model: 
\begin{equation}
g_{x_{1:i}}(x_{1:i}) = \sum_{k=0}^{K-1} \theta_{k+1} \cdot x_{i-k} + \theta_0,
\label{eq:arma}
\end{equation}
where $\theta_k \in \mathbb{R}^{N \times N}, k=1,\dots,K$ and $\theta_0 \in \mathbb{R}^{N}$.
AR model is a generalization of linear model to temporal modeling and thus admits a similar analytical solution. The choice of Markov horizon $K$ makes this model flexible and should be informed by the application at hand.
% The past history horizon $K$ allows flexibility of Markov order enables explanation over customized period
%By manipulate the Markov order, we can tailor the interpretation toward desirable period of events.

%\subsection{Explicit Interpretability Game}
\label{sec:meta}
{\bf Explicit Interpretability Game.}
In some cases, we wish to articulate interpretable parts explicitly in the predictor $f$. For example, if we view the predictor as approximately locally linear, we could explicitly parameterize $f$ in a manner that highlights these linear coefficients. To this end, in the temporal domain, we can explicate the locally linear assumption already in the parametrization of $\mu$:
\begin{align}
\mu(x_{1:i}) & = \sum_{k=0}^{K-1} \hat{\theta}(x_{1:i})_{k+1} \cdot x_{i-k} + \hat{\theta}_0(x_{1:i}),\label{eq:explicit}
\end{align}
which we can write as $\hat{\theta}_\text{AR}(x_{1:i}) + \hat{\theta}_0(x_{1:i})$, where $\hat{\theta}$ and $\hat{\theta}_0$ are learned as recurrent networks. However, this explicit parameterization is relevant only if we further encourage them to act their part, i.e., that the locally linear part of $\mu(x_{1:i})$ is really expressed by $\hat{\theta}(x_{1:i})_{k}$, $k=1,\ldots,K$.

%However, in order to effectively establish the interpretability of $\hat{\theta}$ and $\hat{\theta}_0$ we must address some subtle points. We note that if $\hat{\theta}_0$ is parametrized by non-parametric function\todo{seemingly contradictory terms! parametrized, non-parametric}, then all the $\hat{\theta}$ can be ignored, then there is no valid interpretability in the form of AR parameters in (\ref{eq:explicit}).

To this end, we formulate a refined game that defines the discrepancy measure for the explainer in a coefficient specific manner, separately for $\hat{\theta}_\text{AR}$ and $\hat{\theta}_0$ so as to locally mimic the AR and constant family, respectively. % \emph{without offset} $\mathcal{G_\text{AR}}$ and constant family $\mathcal{G_C}$.
The objective of local explainer at $x_{1:i}$ with respect to $\hat{\theta}_\text{AR}$ and $\hat{\theta}_0$ then becomes
\begin{align}
& \min_{g_{x_{1:i}} \in \mathcal{G_\text{AR}}} 
\frac{1}{2\epsilon+1} 
\sum_{i'=i-\epsilon}^{i+\epsilon} 
%d(\sum_{k=0}^{K-1} \hat{\theta}(x_{1:i'})_{k+1} \cdot x_{i'-k}, g_{x_{1:i}}(x_{1:i'}))\nonumber\\
\|\hat{\theta}_\text{AR}(x_{1:i'}), g_{x_{1:i}}(x_{1:i'})\|_2^2\nonumber\\
& + \min_{\bar{g}_{x_{1:i}} \in \mathcal{G_C}} 
\frac{1}{2\epsilon+1} 
\sum_{i'=i-\epsilon}^{i+\epsilon} 
\|\hat{\theta}_{0}(x_{1:i'}), \bar{g}_{x_{1:i}}\|_2^2,
\label{eq:factor_local_deviation}
\end{align}
where $\mathcal{G}_\text{AR}$ is the family of AR models (which does not include any offset/bias, consistent with $\hat{\theta}_\text{AR }$) and $\mathcal{G_C}$ is simply the set of constant vectors. The objective for $f$ is defined analogously, symmetrically or asymmetrically. For simplicity, our notation doesn't include end-point boundary cases with respect to the neighborhoods. 

%% file: experiment.tex
\section{Experiments}
\label{sec:experiment}

\begin{table}[t]
%\vspace{2mm}
\centering
\begin{tabular}{ l  c  c  c }  
\hline
\bf Dataset & \bf\small \# of seq. & \small \bf input len. & \small \bf output len. \\
\hline\hline
Stock & 15,912 & 30 & 7\\
Bearing & 200,736 & 80 & 20\\
\hline
\end{tabular}
\vspace{-2mm}
\caption{Dataset statistics}
\vspace{-2mm}
\label{tab:stat}
\end{table}	
\begin{table}[t]
%\vspace{2mm}
\centering
\begin{tabular}{ l  c  c  c }  
\hline
\bf Stock & \bf\small Error & \small \bf Deviation & \small \bf TV\\
\hline\hline
AR & 1.557 & 0.000 & 0.000\\
\hline
game-implicit & 1.478 & 0.427 & 0.000\\
deep-implicit & 1.472 & 0.571 & 0.000\\
\hline
game-explicit & 1.479 & 0.531 & 73.745\\
deep-explicit & 1.475 & 0.754 & 91.664\\
\hline
\hline
\bf Bearing & \bf\small Error & \small \bf Deviation & \small \bf TV\\
\hline\hline
AR & 9.832 & 0.000 & 0.000\\
\hline
game-implicit & 8.309 & 3.431 & 5.706\\
deep-implicit & 8.136 & 4.197 & 7.341\\
\hline
game-explicit & 8.307 & 4.177 & 27.533\\
deep-explicit & 8.151 & 6.134 & 29.756\\
\hline
\end{tabular}
\vspace{-2mm}
\caption{Performance. All units are in $10^{-2}$.}
\vspace{-6mm}
\label{tab:peref}
\end{table}	

\iffalse
\begin{table}[t]
%\vspace{2mm}
\centering
\begin{tabular}{ l  c  c  c }  
\hline
\bf Stock & \bf\small Error & \small \bf Deviation & \small \bf TV\\
\hline\hline
AR & 0.01557 & 0.00000 & 0.00000\\
\hline
game-implicit & 0.01478 & 0.00427 & 0.00000\\
deep-implicit & 0.01472 & 0.00571 & 0.00000\\
\hline
game-explicit & 0.01479 & 0.00531 & 0.73745\\
deep-explicit & 0.01475 & 0.00754 & 0.91664\\
\hline
\hline
\bf Bearing & \bf\small Error & \small \bf Deviation & \small \bf TV\\
\hline\hline
AR & 0.09832 & 0.00000 & 0.00000\\
\hline
game-implicit & 0.08309 & 0.03431 & 0.05706\\
deep-implicit & 0.08136 & 0.04197 & 0.07341\\
\hline
game-explicit & 0.08307 & 0.04177 & 0.27533\\
deep-explicit & 0.08151 & 0.06134 & 0.29756\\
\hline
\end{tabular}
\vspace{-2mm}
\caption{Performance}
\vspace{-6mm}
\label{tab:peref}
\end{table}	
\fi

\begin{table}[t]
%\vspace{2mm}
\centering
\begin{tabular}{ l c  c  c  c c}  
\hline
{$\times 10^{-2}$} & \bf\small $\lambda=0.$ & \bf\small $\lambda=0.1$ & \small \bf $\lambda=1$ & \small \bf $\lambda=10$ & AR\\
\hline\hline
Error & 8.136 & 8.057 & 8.309 & 9.284 & 9.832\\
Deviation & 4.197 & 4.178 & 3.431 & 1.127 & 0.000\\
TV & 7.341 & 7.197 & 5.706 & 1.177 & 0.000\\
\hline
\end{tabular}
\vspace{-2mm}
\caption{Implicit game on bearing dataset when $\lambda$ varies.}
\vspace{-4.25mm}
\label{tab:compare}
\end{table}	

\iffalse
  \begin{table}[t]
  %\vspace{2mm}
  \centering
  \begin{tabular}{ l c  c  c  c }  
  \hline
  \bf {} & \bf\small $\lambda=0.$ & \bf\small $\lambda=0.1$ & \small \bf $\lambda=1$ & \small \bf $\lambda=10$\\
  \hline\hline
  Error & 0.08136 & 0.08057 & 0.08309 & 0.09284\\
  Deviation & 0.04197 & 0.04178 & 0.03431 & 0.01127\\
  TV & 0.07341 & 0.07197 & 0.05706 & 0.01177 \\
  \hline
  \end{tabular}
  \vspace{-2mm}
  \caption{Implicit game on bearing dataset when $\lambda$ varies.}
  \vspace{-4.mm}
  \label{tab:compare}
  \end{table}	
\fi

\begin{figure}[t]
\centering
  \includegraphics[width=1.1\linewidth]{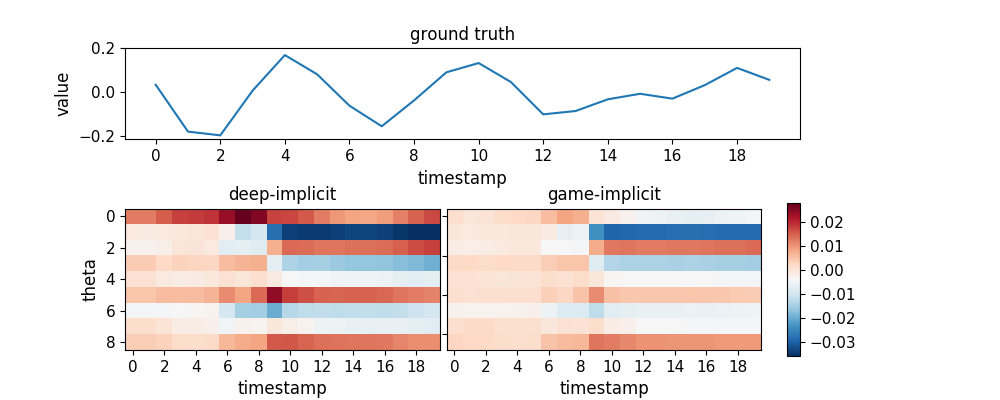}
  \vspace{-8mm}
  \caption{Visualization of weight vectors for predicting the first channel along each autoregressive timestamp in bearing dataset. The $y$-axis from $0$ to $8$ denotes $(\theta_0)_1$, $(\theta_1)_{1, 1:4}$ and $(\theta_2)_{1, 1:4}$}.
  \label{fig:visual}
  \vspace{-9mm}
\end{figure}

We validate our approach on two real-world applications: a bearing dataset from NASA~\cite{bearing_dataset} and a stock market dataset consisting of historical data from the S\&P 500 
%index\footnote{\url{www.kaggle.com/camnugent/sandp500/data}}. 
index\footnote{\url{www.kaggle.com/camnugent/sandp500/data}}. 
The bearing dataset records 4-channel acceleration data on 4 co-located bearings, and stock dataset records daily prices (4 channels in open, high, low, and close).
Due to the diverse range of stock prices, we transform the data to daily percentage difference.
We divide the sequence into disjoint subsequences and train the sequence generative model on them.
The input and output length are decided based on the nature of dataset.
The bearing dataset has a high frequency period of 5 points and low frequency period of 20 points. On the stock dataset, we used 1 month to predict the next week's prices. 
The statistics of the processed dataset is shown in \tabref{tab:stat}.
We randomly sample $85\%$, $5\%$, and $10\%$ of the data for training, validation, and testing.

We set neighborhood $\epsilon$ and Markov order $K$ to be $6$ and $7$ to impose sequence-level coherence in stock dataset; and $9$ and $2$ for smooth variation in bearing dataset.
We parametrize $\mu(\cdot)$ and $\Sigma(\cdot)$ jointly by stacking $1$ layer of CNN, LSTM, and $2$ fully connected layers. % implemented in \texttt{Tensorflow}~\cite{abadi2016tensorflow}. 
We use Ridge regression\footnote{Ridge is used to alleviate degenerate cases of linear regression.} with default parameter in \texttt{scikit-learn}~\cite{scikit-learn} to implement the AR model. For efficiency, $10\%$ of the sequences are sampled for regularization in each batch.

We compare our asymmetric game-theoretic approach (`game') against the same model class without an explainer (`deep'). We use `implicit' label to distinguish predictors from those `explicitly' written in an AR-like form. Evaluation involves three different types of errors: `Error' is the root mean squared error (RMSE) between greedy autoregressive generation and the ground truth; `Deviation' is RMSE between the model prediction $\mu(x_{1:i})$ and the explainer $g_{x_{1:i}}(x_{1:i})$, estimated also for `deep' that is not guided by the explainer; and `TV' is the average total variation of $[\hat{\theta}, \hat{\theta}_0]$ over any two consecutive time points. For testing, the explainer is estimated based on a greedy autoregressive generative trajectory. TV for implicit models is based on the parameters of the explainer $g_{x_t}$. For explicit models, deviation RMSE is the sum of AR and constant deviations as in Eq. (\ref{eq:factor_local_deviation}), thus not directly comparable to the implicit `Deviation' based only on output differences.

The results are shown in \tabref{tab:peref}.
TV is not meaningful for implicit formulation in stock dataset due to the size of the neighborhood.
The proposed game reduces the gap between deep learning model and AR model on deviation and TV, while retaining promising prediction accuracy.

We also show results of implicit setting on bearing dataset for different $\lambda$ in \tabref{tab:compare}. The trends in increasing error and decreasing deviation and TV are quite monotonic with $\lambda$. When $\lambda = 0.1$, the `game' model is even more accurate than `deep' model due to regularization effect of the game.

We visualize the explanations from implicit setting ($[\theta_0, \theta]$ in explainer model) over autoregressive generative trajectories in \figref{fig:visual}. The explanation from the `game' model is more stable. 
Compared to the ground truth, different temporal pattern after the $9^\text{th}$ point is captured by the explanations.
%\vspace{-1mm}

%% file: conclusion.tex
\vspace{-3mm}
\section{Conclusion}
\label{sec:conclusion}
\vspace{-0.5mm}
We provide a novel game-theoretic approach to interpretable temporal modeling. The game articulates how the predictor accuracy can be traded off against locally agreeing with a simple axiomatically interpretable explainer. The work opens up many avenues for future work, from theoretical analysis of the co-operative games to estimation of interpretable unfolded trajectories through GANs.